\documentclass[article]{IEEEtran}

\usepackage{cite}
\usepackage{amsmath,amssymb,amsfonts}
\usepackage{algorithmic}
\usepackage{graphicx}
\usepackage{textcomp}
\usepackage{xcolor}
\usepackage{booktabs}
\usepackage{siunitx}
\usepackage{url}
\usepackage{caption}
\usepackage{glossaries}
\usepackage{longtable}
\usepackage{soul}
\usepackage{caption}

\renewcommand{\thesubsection}{\thesection.\arabic{subsection}}

\makeglossaries
\newglossarystyle{mylong}{
    {\begin{longtable}{lp{\glsdescwidth}}}%
    {\end{longtable}}%
  \renewcommand*{\glsgroupheading}[1]{}%

}
\def\BibTeX{{\rm B\kern-.05em{\sc i\kern-.025em b}\kern-.08em
    T\kern-.1667em\lower.7ex\hbox{E}\kern-.125emX}}

\begin{document}

\title{A temporal scale transformer framework for precise remaining useful life prediction in fuel cells
}



\author{
    \IEEEauthorblockN{
        Zezhi Tang$^{1*}$,
        Xiaoyu Chen$^2$,
        Xin Jin$^2$,
        Benyuan Zhang$^2$,
        Wenyu Liang$^2$,\\
        Yi Dong$^3$,
        George Panoutsos$^2$,
        Zepeng Liu$^4$
        Mingyang Lu$^5$
    }

\thanks{$^1$ Department of Computer Science, University College London.}
\thanks{$^2$ Department of Automatic Control and Systems Engineering, University of Sheffield.}
\thanks{$^3$ Department of Computer Science, University of Liverpool.}
\thanks{$^4$ School of Engineering, Newcastle University.}
\thanks{$^5$ Center for Nondestructive Evaluation, Iowa State University.}
\thanks{Zezhi Tang, Xiaoyu Chen, Xin Jin, Benyuan Zhang and Wenyu Liang contributed equally to this work.}
\thanks{Correspondence: zezhi.tang@ucl.ac.uk}
}

\maketitle

\begin{abstract}

In the task of exploring Predictive Health Management (\gls{PHM}) strategies for Proton Exchange Membrane Fuel Cells (\gls{PEMFC}), the Transformer model, due to its outstanding performance in numerous fields, has been widely applied within data-driven models. However, for time series analysis tasks, using the self-attention mechanism in the Transformer model results in a complexity of the input sequence squared, often leading to low computational efficiency. Additionally, the model faces the challenge of effectively extracting both global long-term dependencies and local detail features simultaneously. To address these issues, this paper proposes a novel solution in the form of an improved version of the inverted Transformer (\gls{iTransformer}) - the Temporal Scale Transformer (\gls{TSTransformer}). Specifically, unlike traditional Transformers, which model each timestep as an input token, the new embedding method maps sequences of different lengths into tokens at different stages for inter-sequence modelling, allowing multivariate correlations to be described through the attention mechanism and sequence representations to be encoded through feed forward networks(\gls{FFN}). Notably, incorporating a one-dimensional convolutional layer into the multivariate attention at different stages for multi-level scaling of the K and V matrices not only enhances the ability to extract local features and capture characteristics at different temporal scales but also reduces the number of tokens and computational costs. This enables the model to effectively capture the ability to extract local features in time series data. Through rigorous experimental research on the model and comparative analysis with established models (such as Long Short-Term Memory (\gls{LSTM}), iTransformer, and Transformer). The results collectively highlight the potential of the TSTransformer model as a powerful tool for advancing PHM in the renewable energy sector and effectively solving the problems associated with pure Transformer backbone models in data-driven time series tasks.
\end{abstract}


\newglossaryentry{PHM}{
    name={PHM},
    description={prognostic health management}
}
\newglossaryentry{PEMFC}{
    name={PEMFC},
    description={proton exchange membrane fuel cell}
}
\newglossaryentry{RUL}{
    name={RUL},
    description={remaining useful life}
}

\newglossaryentry{RMSE}{
    name={RMSE},
    description={root mean square error}
}
\newglossaryentry{HI}{
    name={HI},
    description={health indicator}
}
\newglossaryentry{RE}{
    name={RE},
    description={relative error}
}

\newglossaryentry{LSTM}{
    name={LSTM},
    description={long short-term memory}
}
\newglossaryentry{RNN}{
    name={RNN},
    description={recurrent neural network}
}
\newglossaryentry{CNN}{
    name={CNN},
    description={convolutional neural networks}
}

\newglossaryentry{NLP}{
    name={NLP},
    description={natural language processing}
}

\newglossaryentry{MEA}{
    name={MEA},
    description={membrane electrode assembly}
}
\newglossaryentry{MLP}{
    name={MLP},
    description={Multi-Layer Perceptron}
}
\newglossaryentry{CV}{
    name={CV},
    description={computer vision}
}
\newglossaryentry{MSA}{
    name={MSA},
    description={multi-scaled attention}
}
\newglossaryentry{FFN}{
    name={FFN},
    description={feed forward network}
}
\newglossaryentry{LOESS}{
    name={LOESS},
    description={Local Weighted Scatterplot Smoothing}
}
\newglossaryentry{Bi-GRU}{
    name={Bi-GRU},
    description={Bidirectional Gated Recurrent Units}
}
\newglossaryentry{Bi-LSTM}{
    name={Bi-LSTM},
    description={Bidirectional Long Short-Term Memory Networks}
}
\newglossaryentry{ESN}{
    name={ESN},
    description={Echo State Networks}
}
\newglossaryentry{MK-RVM}{
    name={MK-RVM},
    description={multi-kernel relevance vector machine}
}
\newglossaryentry{EMD}{
    name={EMD},
    description={Empirical Mode Decomposition}
}
\newglossaryentry{VAE-DGP}{
    name={VAE-DGP},
    description={variational auto-encoded deep Gaussian
process}
}
\newglossaryentry{TSTransformer}{
    name={TSTransformer},
    description={temporal scale transformer}
}
\newglossaryentry{iTransformer}{
    name={iTransformer},
    description={inverted transformer}
}


\noindent \textbf{Notation}{\vspace{2pt}}

\noindent 
 \begin{tabular}{@{}ll}
PHM & Prognostic Health Management\\
PEMFC & Proton Exchange Membrane Fuel Cell\\
RUL & Remaining Useful Life\\
RMSE & Root Mean Square Error\\
HI & Health Indicator\\
RE & Relative Error\\
LSTM & Long Short-Term Memory\\
RNN & Recurrent Neural Network\\
  \end{tabular}


\noindent
 \begin{tabular}{@{}ll}
CNN & Convolutional Neural Networks\\
NLP & Natural Language Processing\\
MLP & Multi-Layer Perceptron\\
CV & Computer Vision\\
MSA & Multi-Scaled Attention\\
FFN & Feed Forward Network\\
LOESS & Local Weighted Scatterplot Smoothing\\
Bi-GRU & Bidirectional Gated Recurrent Units\\
Bi-LSTM & Bidirectional Long Short-Term Memory Networks\\
ESN & Echo State Networks\\
MK-RVM & Multi-Kernel Relevance Vector Machine\\
EMD & Empirical Mode Decomposition\\
VAE-DGP & Variational Auto-Encoded Deep Gaussian Process\\
TSTransformer & Temporal Scale Transformer\\
iTransformer & inverted Transformer

 \end{tabular}


\section{Introduction}
Energy consumption is expected to rise in the following decades, underscoring the critical need for renewable energy sources that significantly lower carbon emissions. The \gls{PEMFC} is a type of hydrogen fuel cell that has gained popularity due to its high efficiency, fast response time, and clean waste. It can be applied to different areas, including industrial use, portable electronics, and alternative transportation power sources \cite{cai2020proton}. Nonetheless, the deterioration of vital components shortens the lifespan of the \gls{PEMFC}, a challenge that necessitates advanced monitoring and prognostic strategies.

In this paper, \gls{PHM} systems can estimate the degradation status based on past operational profiles and current data, which have proven essential in determining the degradation state of the \gls{PEMFC} \cite{sutharssan2017review}. \gls{PHM} strategies can be classified into three groups: hybrid, data-driven, and model-driven approaches, and prediction model selection is the central part of the prognostic \cite{hua2022review}. The selection of an efficient health indicator (\gls{HI}) is essential to this prognostic procedure, and the stack voltage is frequently used as a trustworthy \gls{HI} due to its strong association with the health state of the cell \cite{hua2020health}.

The prediction of RUL through model-driven approaches primarily depends on the fuel cell loading conditions, material properties, and degradation and failure mechanisms. It is adaptable to various fuel cell types and does not require a significant amount of experimental data \cite{hua2022review}. For example, Jouin et al. \cite{jouin2014prognostics} proposed a prediction framework based on particle filtering for estimating the \gls{RUL} of \gls{PEMFC}s. This method combines unobservable states (such as degradation level) with a physical model to predict the future behaviour and RUL of PEMFCs through probability distribution. Moreover, \cite{chen2019fuel} presents a predictive method for \gls{PEMFC}s applied to vehicles. The method employs the unscented Kalman filter algorithm combined with a proposed voltage degradation model, using real field data for validation. Jouin et al.\cite{jouin2016degradations} proposed an in-depth literature review and degradation analysis based on \gls{PEMFC} reactors. This analysis defines an appropriate vocabulary and a clear, limited framework for executing predictions. The model was validated on datasets of different health assessment and prediction tasks. Nevertheless, the model-based method is complex, and it is challenging to accurately establish degradation models for the \gls{PEMFC}. This is due to the fact that the \gls{PEMFC} degradation mechanisms operate on tough multi-time and multi-physical scales \cite{liu2019remaining}. 

The technique that relies on monitored historical data to anticipate the remaining lifespan of the \gls{PEMFC} is a data-driven method. This approach is rapid and efficient since it does not require knowledge of fuel cell models or systems \cite{liu2019remaining1}.  Liu et al\cite{liu2019remaining1}. proposed a new method for predicting the \gls{RUL} based on \gls{LSTM} Recurrent Neural Networks. The outstanding contribution of this method in the data-driven field is that it can effectively remove noise and spikes through regular interval sampling and Local Weighted Scatterplot Smoothing (\gls{LOESS}) while retaining the main trend of the original data. In \cite{wang2020stacked}, stack \gls{LSTM} is formed by two \gls{LSTM} models that add two dropout parameters and further improve the model performance. Li et al\cite{li2022degradation}. proposed a predictive framework based on the fusion of Bidirectional Long Short-Term Memory Networks (\gls{Bi-LSTM}), Bidirectional Gated Recurrent Units (\gls{Bi-GRU}), and Echo State Networks (\gls{ESN}) for short-term degradation prediction and \gls{RUL} estimation of \gls{PEMFC}. The outstanding contribution of this framework in the data-driven field is that it can achieve accurate predictions with fewer training datasets, and compared with traditional machine learning methods, it shows better predictive performance. Peng et al\cite{peng2023remaining}. proposed a prediction method based on CNN and LSTM. Their method uses the Savitzky Golay smoothing dataset for data processing, removes outliers using box plots, and standardizes the dataset using a Z-score. This study has achieved significant results in the universality testing of long-term and short-term predictions. Nonetheless, Artificial intelligence technology drives the popularity of data-driven prediction models, but collecting enough data for training is challenging.

A hybrid model is a statistical model that combines different distribution characteristics, typically consisting of two or more components, and can better capture complex patterns and relationships in data. Tian et al\cite{tian2023remaining}. proposed a hybrid model of a multi-kernel relevance vector machine (\gls{MK-RVM}) based on a voltage recovery model and Bayesian optimization. Preprocess the data using the Empirical Mode Decomposition (\gls{EMD}) method and train the model using \gls{MK-RVM}. Using the Bayesian optimization algorithm to optimize the weight coefficients of the kernel function, the parameters of the prediction model are updated, and the voltage recovery model is added to the prediction model to achieve a fast and accurate prediction of PEMFC RUL. Liu et al\cite{liu2019remaining}. proposed a hybrid prediction method. In the first stage, an automatic machine learning algorithm based on evolutionary algorithms and adaptive neuro-fuzzy inference systems is used to predict long-term degradation trends. The second stage is based on the degradation data obtained in the first stage, and RUL estimation is achieved through a semi-empirical degradation model and the proposed adaptive unscented Kalman filtering algorithm. In addition, this method also achieves automatic parameter adjustment through a particle swarm optimization algorithm, improving the accuracy and convenience of prediction. However, hybrid models also have some drawbacks. For example, the complexity and computational complexity of the model results in the need for more computing resources and time in practical applications. 

Given the computational complexity and the requisite extensive domain knowledge inherent in model-driven and hybrid-driven approaches, there is an increasing shift towards data-driven methodologies for \gls{PEMFC} \gls{RUL} prediction. These data-driven methods capitalize on directly extracting insights from operational data, circumventing the complexities associated with traditional models. However, ensuring prediction accuracy and obtaining high-quality data is still the goal of this direction's efforts. In the realm of \gls{PEMFC} prognostics, traditional data-driven models, including Gaussian process regression and \gls{LSTM}-based frameworks, have made strides but also faced limitations in capturing the complex dynamics of fuel cell degradation. For instance, \gls{LSTM} models, despite their proficiency in temporal data modelling, may need to fully grasp the intricacies of multivariate time series characteristic of \gls{PEMFC} systems \cite{deng2022degradation, zhang2023evolutionary}.
The recent surge in interest towards the innovation of Transformer \cite{vaswani2017attention} in the realm of deep learning is largely due to its superior performances across various fields, such as natural language processing (\gls{NLP}) \cite{han2021pre}, computer vision(\gls{CV}) \cite{khan2022transformers}. In the last few years, a myriad of Transformers have been introduced and have significantly elevated the standard performance of diverse tasks. A novel Transformer-based \gls{PEMFC} prognostic framework for predicting long-term degradation is employed in \cite{lv2023transformer}. The introduction of a series-attention mechanism is proposed to enhance global feature representation and overall long-term prognostic performance. showing potential for improvement. In addition, Tang et al. \cite{tang2023novel} presented an innovative end-to-end approach to estimating the \gls{RUL} of the \gls{PEMFC} by fusing the Transformer model with transfer learning. The fundamental principle of this approach is to pre-train a transformer model on a static dataset and then relocate the model to the target dynamic dataset for optimisation. The experimental results show the proposed method has the lowest \gls{RMSE} compared with DNN, \gls{LSTM}, and variational auto-encoded deep Gaussian process(\gls{VAE-DGP}). 

Transformers have proved highly effective in modelling long-range dependencies and interactions within sequential data, making them attractive for time series modelling. Numerous Transformer variants have been proposed to tackle unique challenges in time series modelling. They have been successfully implemented in a range of time series tasks, including but not limited to forecasting \cite{li2019enhancing} \cite{zhou2022fedformer}, anomaly detection \cite{xu2021anomaly} \cite{tuli2022tranad}, and classification \cite{zerveas2021transformer} \cite{yang2021voice2series}. In particular, the seasonality or periodicity is a key feature of time series \cite{wen2021robustperiod}, yet effectively modelling both long-range and short-range temporal dependency while capturing seasonality simultaneously presents an ongoing challenge \cite{wu2020lite}. This recognition of limitations within the conventional Transformer framework inspired the development of the iTransformer\cite{liu2023itransformer} model.
The iTransformer considers sequences from individual channels as singular tokens, embedding them along the sequence dimension while applying attention across the channel dimension. This approach provides a highly adaptable framework for processing time series data, enabling the Transformer to construct bespoke models for different time series data, yielding impressive predictive performance.

Building on the foundation laid by the iTransformer, this paper proposes the TSTransformer, aptly named to reflect its focus on the temporal aspects of PEMFC degradation. This innovative model leverages the core strengths of the iTransformer, emphasizing its capability to embed multivariate time series data as tokens and apply a refined self-attention mechanism. This approach ensures the model captures intricate temporal relationships and correlations across multiple variates, which is vital for precise RUL forecasting. Furthermore, integrating one-dimensional convolutions represents a strategic enhancement, streamlining the attention mechanism's capacity to process extensive time series data efficiently. Such adaptations render the Temporal Scale Transformer particularly effective for long-term RUL estimations, signifying a significant advancement in the model's design to cater to the specific challenges of PEMFC prognostics.
Building on the requirement for robust \gls{PHM} systems to improve the robustness and reliability of the \gls{PEMFC}, this study makes three contributions to improve prognostication in the renewable energy industry, which include:

\begin{itemize}
    \item The optimized Transformer processes multivariate time series data as individual tokens, improving its ability to capture complex temporal dynamics and enhance RUL forecasting accuracy in PEMFC systems.
    
    \item Incorporating one-dimensional convolutions reduces the complexity of key and value matrices, boosting the model’s efficiency and accuracy in long-term RUL predictions for PEMFC applications.

    \item The proposed model is extended to practical experimentation, seamlessly integrating data reconstruction and noise reduction. The experiment yields highly effective outcomes, further demonstrating the performance of the model. 
    
\end{itemize}

\section{Methodology}
\subsection{PEMFC Stacks and Ageing Test}
The two main parts of a single \gls{PEMFC} are the bipolar plates and the membrane electrode assembly (\gls{MEA}). Bipolar plates are essential to fuel cells for temperature control, mechanical support, electrical current collection, gas delivery (hydrogen in the anode and oxygen in the cathode), and mechanical support. The \gls{MEA} has three
distinct parts: the diffusion layer, which is a porous segment of the electrode, the catalyst layer (the reaction zone), and the membrane itself.
These \gls{MEA} component pieces allow for the effective transport of reactants, electrochemical reactions, and fuel cell energy production \cite{gao2013proton}.

\begin{figure}[ht]
  \centering
  \includegraphics[width=0.4\textwidth]{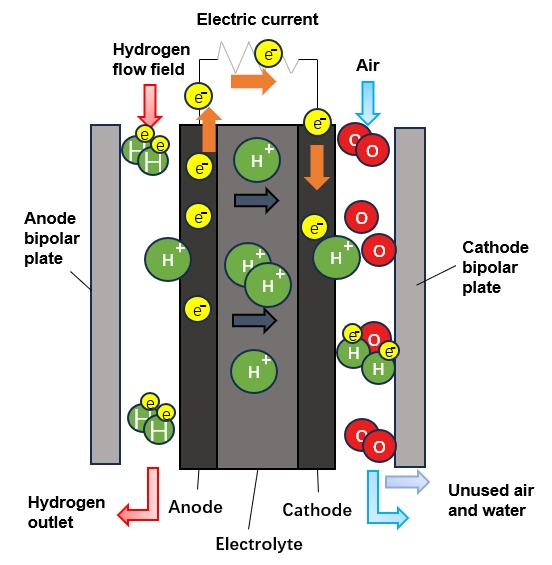}
  \caption{Mechanism structure of a single PEMFC.}
  \label{fig:select feature}
\end{figure}

The degradation of the \gls{PEMFC} can be separated into mechanical degradation and chemical degradation. Non-uniform mechanical stresses lead to mechanical deterioration. For example, variations in temperature and relative humidity can affect the physical dimensions of fuel cell components. On the other hand, chemical deterioration is caused by external disturbances and the ageing process of materials. Moreover, material ageing specifically refers to the polymer breakdown, catalyst dissolution, and bipolar plate corrosion that occur in \gls{MEA} \cite{hua2022review}. The reactants and external conditions are not distributed equally within a single cell surface. Different current densities may occur from this heterogeneity in various parts of the fuel cell system, which could cause different rates of degradation. The degradation states can be observed by conducting the ageing test.

The ageing data used in this article is sourced from the IEEE PHM 2014 Data Challenge quasi-dynamic dataset \cite{gouriveau2014ieee}. The experimental data collection process involved the utilisation of a specialised test bench designed for Fuel Cells with a power capacity of up to 1 kW. The test bench facilitated precise control and measurement of various physical parameters within the FC stack. Parameters listed in Table \ref{tab:controlpara} were monitored and controlled during the experiments, and the Labview interface enabled real-time monitoring of single cell and stack voltages, as well as current. 

\begin{table}[ht]
\centering
\caption{Controllable range of eight physical parameters}
\label{tab:controlpara}
\begin{tabular}{ll}
\hline
\textbf{Parameters} & \textbf{Range of control} \\
\hline
Cooling temperature & 20-80 (°C) \\
Cooling flow & 0-10 (l/min) \\
Gas temperature & 20-80 (°C) \\
Gas humidification & 0-100 (\%RH) \\
Air flow & 0-100 (l/min) \\
H2 flow & 0-30 (l/min) \\
Gas pressure & 0-2 (bars) \\
Fuel Cell current & 0-300 (A) \\
\hline
\end{tabular}
\end{table}

The FC stacks under investigation were 5-cell stacks assembled at FCLAB\cite{gouriveau2014ieee}, each cell featuring a 100 cm\textsuperscript{2} active area. The nominal current density of a \gls{PEMFC} was set at 0.70 A/cm\textsuperscript{2}, with a maximum current density of 1 A/cm\textsuperscript{2}. The experimental setup included two ageing tests: a stationary regime test conducted under nominal operating conditions (FC1), serving as a reference, and a dynamic current testing condition with high-frequency current ripples (FC2). 
The ageing data provided to challenge participants included monitoring data detailing power loads, temperatures, hydrogen and air stoichiometry rates, among other parameters. These parameters, presented in Table\ref{tab:features}, facilitated the assessment of normal or accelerated ageing of the fuel cell stacks, with a focus on degradation phenomena captured through voltage drop over time. 

\begin{table}[ht]
\centering
\caption{Health monitoring features}
\label{tab:features}
\begin{tabular}{ll}
\hline
\textbf{Features} & \textbf{Explanations} \\
\hline
Time (h) & Ageing time  \\
U1-U5, Utot (V) &  Five single cell voltage and stack voltage \\
I (A), J (A/cm\textsuperscript{2}) & Current and current density  \\
TinH2, ToutH2 (°C) & The inlet and outlet hydrogen gas temp. \\
TinAIR, ToutAIR (°C) & The Inlet and outlet air temp. \\
TinWAT, ToutWAT (°C) & The Inlet and outlet cooling water temp. \\
PinH2, PoutH2 (mbara) & The inlet and outlet hydrogen gas pressure \\
PinAIR, PoutAIR (mbara) & The inlet and outlet air pressure \\
DinH2, DoutH2 (L/min) & The inlet and outlet hydrogen gas flow rate \\
DinAIR, DoutAIR (L/min) & The inlet and outlet air flow rate \\
DWAT (L/min) & The cooling water flow rate \\
HrAIRFC (\%) & The estimated hygrometry of inlet air \\
\hline
\end{tabular}
\end{table}
\subsection{Data preprocessing}
\begin{figure} 
  \centering
  \includegraphics[width=0.5\textwidth]{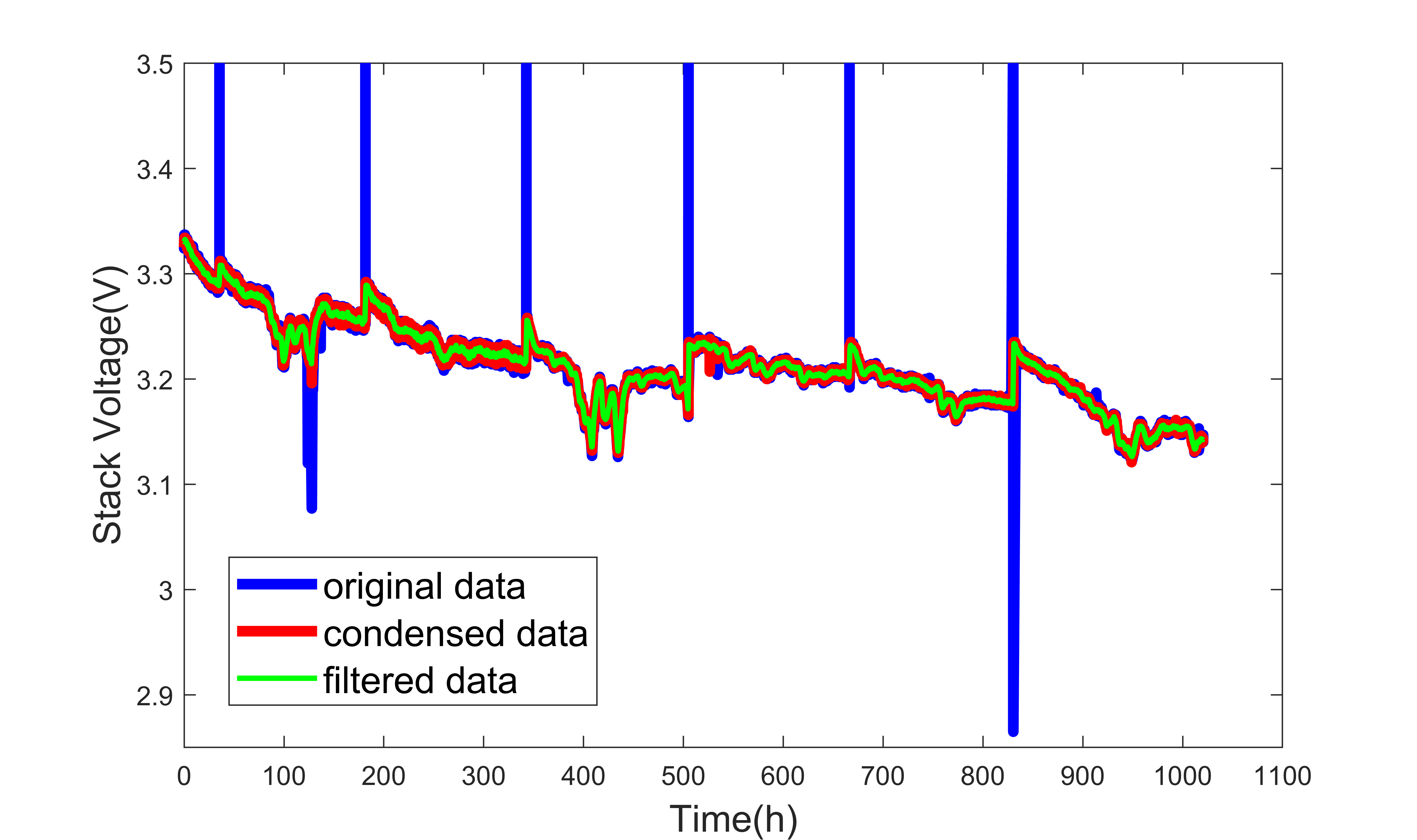}
  \caption{Visualisation of original data, condensed data and filtered data.}
  \label{fig:data preprocessing}
\end{figure}

The durability test collects 127,370 data points from 0 to 1020h. The accuracy of \gls{PHM} systems in determining the \gls{RUL} of the \gls{PEMFC} depends on the efficient preprocessing of time-series data. Due to the significant size of these datasets, the data is condensed by collecting a data point every six minutes, which results in 10,134 data points. Then the moving average filter \cite{smith2013digital} is applied to the condensed dataset to reduce the noise while preserving data primitivity. The original data, condensed data and filtered data are shown in Fig.\ref{fig:data preprocessing}.

\begin{figure*} [!t]
  \centering
  \includegraphics[width=1\textwidth]{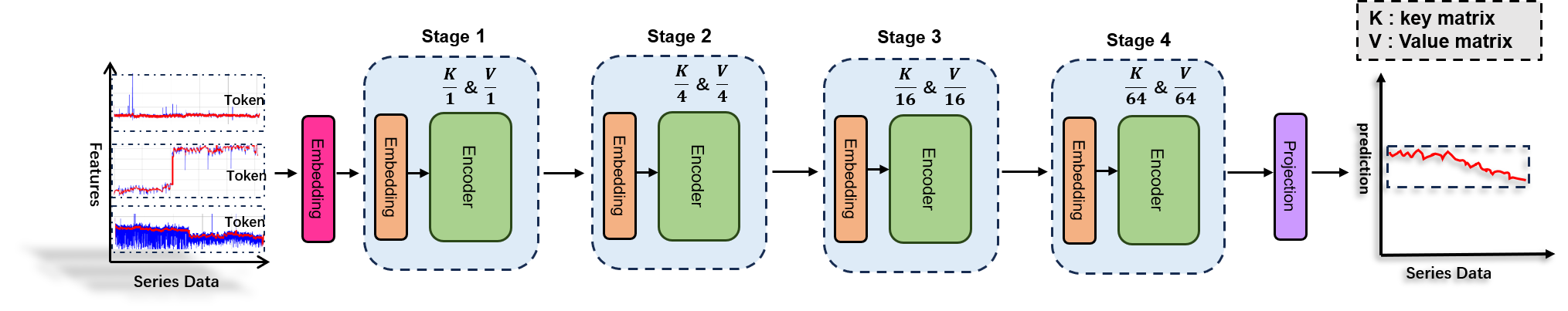}
  \caption{Overall architecture of Temporal scale Transformer(TSTransformer).}
  \label{fig:stage}
\end{figure*}

\subsection{Enhanced Time Series Forecasting Methodology}

In the specified forecasting framework, a dataset \(\mathcal{D}\) comprises a sequence of multivariate observations \(\mathcal{O} = \{\mathbf{o}_1, \ldots, \mathbf{o}_T\}\), with each \(\mathbf{o}_t\) belonging to the feature space \(\mathbb{R}^{M}\) for \(t = 1, \ldots, T\). The term \(T\) denotes the total number of time steps, while \(M\) indicates the count of features per observation, inclusive of the target attribute for prediction. The aim is to extend the historical observations \(\mathcal{O}\) into a future forecast over a horizon \(S\), generating a prediction set \(\mathcal{P} = \{\mathbf{p}_{T+1}, \ldots, \mathbf{p}_{T+S}\} \subseteq \mathbb{R}^{S \times M}\).

The predictive function is designed to process a given input sequence and output a series of future values. The input sequence is first cast into a tensor format amenable to the model. Subsequent predictions are made iteratively, spaced by a defined interval, and the results are aggregated to form the final forecast set \(\mathcal{P}\).

Each \(\mathcal{D}_{t}\) represents the feature vector at a specific time step \(t\), and \(\mathcal{D}^{(m)}\) corresponds to the complete time series for the \(m^{th}\) feature. The model is structured to manage any inconsistencies in \(\mathcal{D}_{t}\) due to non-uniform data recording intervals, applying a uniform statistical treatment to each series \(\mathcal{D}^{(m)}\) to ensure the reliability of the forecast outputs.

\subsection{Inverted Transformer } 

In the methodology of time series forecasting, the iTransformer model brings a new perspective to the analysis of multivariate time series through thoughtful improvements to the traditional Transformer architecture\cite{liu2023itransformer}. The core transformation lies in reconfiguring and optimising various model components to better adapt to the characteristics of time series data.

The Embedding Layer maps each observation in the time series to a high-dimensional representational token. The architecture employs a Multi-Layer Perceptron (MLP) to embed the entire series, thereby addressing the constraints of conventional Transformers, especially in terms of positional embedding. The initial hidden state for each feature \( f \) within the series is formalised as:
\begin{equation}
Z^0_f = \text{Embedding}(\mathcal{O}_{:,f})
\end{equation}
where \( \mathcal{O}_{:,f} \) encapsulates all observation points for feature \( f \), and \( Z^0_f \) is the initial hidden state post-embedding.

Transformation within the architecture is facilitated by a component known as TrmBlock, which merges self-attention with feed-forward mechanisms. This component acts on the entire embedded sequence, enabling the capture of extensive temporal dependencies:
\begin{equation}
Z^{i+1} = \text{TrmBlock}(Z^{i})
\end{equation}
where \( Z^i \) denotes the output from the \( i^{th} \) layer, and \( i \) ranges from 0 to \( L-1 \), with \( L \) being the total number of layers.

The attention mechanism processes each feature sequence individually within the multivariate framework. Vanilla self-attention in which the use of the matrices \(Q\), \(K\), and \(V\) are derived to compute attention scores and elucidate inter-feature dependencies as folllow:
\begin{equation}
attn\_score = \text{softmax}\left(\frac{QK^\intercal}{\sqrt{d_k}}\right)
\end{equation}
\begin{equation}
attention = attn\_score V,
\end{equation}
where \( attn\_score \) is the attention score matrix, with the output being a weighted summation of the values matrix \( V \).
This research proposes a self attention calculation method for PEMFC task optimization, and the innovative method will be discussed in detail in the next paragraph.

Modifications to the \gls{FFN} are two-fold: It is applied per feature token to enhance feature-specific time series representation and utilises the Universal Approximation Theorem to extract intricate patterns:
\begin{equation}
\text{FFN}(Z) = \text{ReLU}(Z W^1 + b^1) 
\end{equation}
where \( W^1 \) is weights, with \( b^1 \) as biases. \( Z \) represents the hidden state.

layer normalization is applied to the series representation of individual variates, taking into account the entire sequence for each feature independently. This helps in handling non-stationarity and measurement inconsistencies more effectively by normalizing across time steps for each feature. The formula used in the iTransformer is:
\begin{equation}
\text{LayerNorm}(Z) = \frac{Z - \mu(Z)}{\sigma(Z)}
\end{equation}
Here, \( Z \) represents the hidden state, \( \mu(Z) \) is the mean, and \( \sigma(Z) \) is the standard deviation of the hidden state elements.

The final predictive output of the architecture is obtained through a projection layer that translates the last layer's output into future series predictions:
\begin{equation}
\hat{\mathcal{P}}_{.,f} = \text{Projection}(Z^{L}_f)
\end{equation}
Here, \( \hat{\mathcal{P}}_{.,f} \) denotes the predicted future values for feature \( f \), with \( Z^{L}_f \) being the final output of the layer.
These improvements of the iTransformer model, especially in applying layer normalisation and \gls{FFN}, significantly enhance the ability of the Transformer model to process time series, thereby providing a more accurate analytical tool for RUL prediction in \gls{PEMFC} systems.
\subsection{Temporal Scale Transformer }

The TSTransformer is developed based on the mechanism of the iTransformer model for temporal data. The TSTransformer represents a paradigm shift in time series analysis, specifically engineered to enhance the prediction of \gls{RUL} in the \gls{PEMFC}. It transcends the conventional self-attention framework by leveraging a multi-scaled approach to process the key (\(K\)) and value (\(V\)) matrices, thereby adeptly capturing temporal dynamics over various scales.

\begin{figure*} [!t]
  \centering
  \includegraphics[width=1\textwidth]{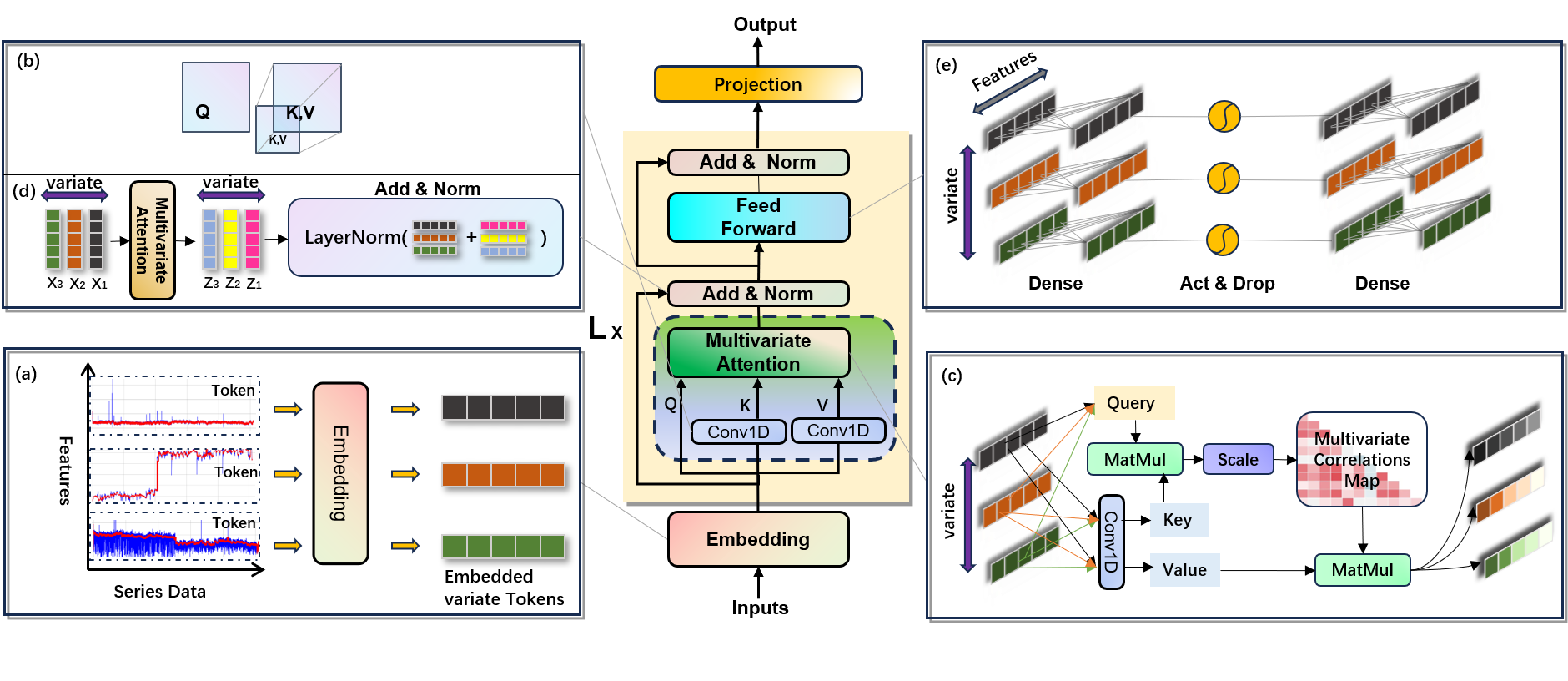}
  \caption{Condensed Overview of the TSTransformer:
(a) Initiates with embedding, transforming time series data into discrete, variate-specific tokens. (b) Implements a convolution-based attention mechanism, adjusting the scale of K and V matrices for diverse temporal resolutions. (c) Applies self-attention to the tokens, highlighting inter-variate correlations and enhancing interpretability. (d) This module uses residual connections and normalisation to harmonise outputs from the attention and feed-forward layers. This ensures a balanced and consistent representation of variables. (e) It concludes with a shared FFN that refines token representations for succinct forecasting.}
  \label{fig:transformer structure}
\end{figure*}

Unlike the vanilla global self-attention used by the standard iTransformer, which processes time series at a single scale, TSTransformer introduces multiple levels of scaling of the \(K\) and \(V\) matrices at different stages, as illustrated in the provided schematic in Fig. \ref{fig:stage} in the provided schematic. By applying one-dimensional deep convolution to the \(K\) and \(V\) matrices at different stages of the self-attention module for spatial reduction, the TST Transformer can recognize complex temporal patterns, encompassing a wide range of trends and nuances in time series data, while reducing the number of markers and computational cost. We believe this is due to the different scales of convolution of the \(K\) and \(V\) matrices, which allows the model to focus on features at different scales on different stages, thus learning both the trend of the time series as a whole and focusing on a wide range of details in the time series at the same time, and reduces the complexity of the computation.

\begin{enumerate}
    \item \textbf{Token Embedding and 1D-Convolution(Conv1D) for K/V}:
    As illustrated in the Embedding process ( Fig.\ref{fig:transformer structure}(a)), each input token \(X\) is embedded through three different linear layers to obtain the three key matrices required by the self-attention module: q k v. We then spatially reduce k/v through two independent 1-D depth-convolutional layers (\text{Conv1D}), with each stage \(i\) scaled by \(R_i\):
    \begin{equation}
      K_i, V_i = \text{Conv1D}(X, R_i)
    \end{equation}
    Here, \(K_i\) and \(V_i\) correspond to the k/v scaling ratio of the ith stage, respectively, as shown by the convolution markers in Fig. \ref{fig:transformer structure}(b). The scaling ratios \(R_i\) are calibrated as \(R_1 = 1\), \(R_2 = 2^{-2}\), \(R_3 = 2^{-4}\), and \(R_4 = 2^{-5}\), with respect to the different scaling of the spatial dimension.

    \item \textbf{Multi-Scaled Attention (MSA)}:
    \gls{MSA}, as portrayed in Fig.\ref{fig:transformer structure}(c), is computed using the query matrix (\(Q\)) at the original scale and the spatially reduced keys (\(K_i\)) and values (\(V_i\)):
    \begin{equation}
      attention_i = \text{softmax}\left(\frac{Q {K_i}^T }{\sqrt{d_{k_i}}}\right)V_i 
    \end{equation}
    In this formulation, \(Q\), \(K_i\), and \(V_i\) are matrices representing queries, keys, and values, respectively, with \(attention_i\) being the attention score matrix for stage \(i\). The term \(\sqrt{d_{k_i}}\) serves as a scaling factor to normalise the dot products in the attention computation, where \(d_{k_i}\) is the dimensionality of the key vectors at stage \(i\), ensuring that the softmax function operates over a range of values that leads to a stable gradient during training.

    \item \textbf{Attention Output with Residual Connection}:
    The attention output \(O\), integrating information across the various scales with a residual connection to the original input token, is conceptually shown in Fig.\ref{fig:transformer structure}(e):
    \begin{equation}
      O = X + attention_i(Q, K_i, V_i)
    \end{equation}
    Here, \(O\) represents the aggregated output, \(X\) denotes the initial input token matrix, and \(attention_i(Q, K_i, V_i)\) encapsulates the stage-specific attention outputs, with the summation encompassing all stages, thereby ensuring a balanced and consistent representation for forecasting.
\end{enumerate}

The TSTransformer architecture, with its multi-scale temporal processing, represents a significant enhancement in predictive accuracy for \gls{RUL} in the \gls{PEMFC}. The integration of spatially scaled attention within a robust Transformer framework makes TSTransformer a powerful model for complex time series forecasting tasks.

\subsection{Evaluation Metrics}
In this project, the performance of the prediction model is assessed using key metrics such as \gls{RMSE}, the percentage error of the forecast (\%Er\textsubscript{FT}), and the accuracy score of the \gls{RUL} estimate (A\textsubscript{FT})\cite{mao_jackson_2016}. To compute these metrics, several voltage loss thresholds are defined that signify the degradation states of the fuel cell system. The RUL is recognised as the time before a certain percentage of voltage loss, defined by thresholds such as 3.5\%, 4.0\%, 4.5\%, 5.0\%, and 5.5\%\cite{sun2023improved}. Particularly, the initial voltage of the fuel cell in this study is set at 3.325V, which serves as the basis for determining the voltage loss thresholds. The \gls{RMSE} is used to measure the discrepancy between the actual cell voltages and the predicted values; \%Er\textsubscript{FT} reflects the percentage error in predictions at each voltage loss threshold, while A\textsubscript{FT} is a score given based on the overall performance of all \gls{RUL} estimates.

\textbf{RMSE} is a statistical measure that quantifies the difference between predicted values ($\hat{y}_i$) and actual values ($y_i$), given by the formula:
\begin{equation}
RMSE = \sqrt{\frac{1}{N} \sum_{i=1}^{N} (\hat{y}_i - y_i)^2}
\end{equation}
where $N$ is the number of forecast points.

\textbf{\%Er\textsubscript{FT}} is calculated by comparing the difference between the actual RUL and the forecast RUL, expressed as:
\begin{equation}
\%Er_{FT} = \frac{RUL_{true} - RUL_{prognostic}}{RUL_{true}} \times 100\%
\end{equation}
A positive value indicates an early forecast, whereas a negative value indicates a delayed forecast.

\textbf{ A\textsubscript{FT}} is a function based on \%Er\textsubscript{FT}, which changes form depending on whether the forecast is early (\%Er\textsubscript{FT} \textgreater 0) or late (\%Er\textsubscript{FT} \( \leq \)0). If the forecast is early, the accuracy score decreases according to a certain function; if it is late, it decreases more strictly, reflecting the seriousness of late predictions in practical applications. The accuracy score is calculated as:
\begin{equation}
A_{FT} = 
  \begin{cases} 
   \exp(-\ln(0.5) \cdot (\%Er_{FT}/5)) & \text{if } \%Er_{FT} \leq 0 \\
   \exp(\ln(0.5) \cdot (\%Er_{FT}/20)) & \text{if } \%Er_{FT} > 0
  \end{cases}
\end{equation}

The final \gls{RUL} assessment score ($Score_{RUL}$) is the mean of all A\textsubscript{FT} values, defined as:
\begin{equation}
Score_{RUL} = \frac{1}{5} \sum_{} A_{FT},FT \in \{3.5\%,4\%, 4.5\%, 5\%, 5.5\%\}
\end{equation}
The closer this score is to 1, the better the predictive performance of the model.

The paper aims to provide accurate forecasts for the \gls{RUL} of the fuel cell system under different failure thresholds. These evaluation criteria guide the accuracy of the predictions.

\section{Results and discussion}
\subsection{Environment configuration and hardware selection of the model}
The \gls{PHM} system employing a Transformer-based model is developed using Python as the primary programming language. The system's computational environment is equipped with an Intel(R) Core(TM) i7-12700H CPU, supplemented by a substantial memory capacity of 64 GB. The system operates on Windows 11, graphical computations are handled by an NVIDIA RTX 3070Ti Laptop GPU. The software stack includes PyTorch version 2.1.0 and Transformers library version 2.1.1, both of which are compatible with CUDA 12.1.

\subsection{Dataset selection and partitioning}
In this study, to comprehensively evaluate the performance of the predictive model, two fuel cell datasets, FC1 and FC2, were selected. Specifically, this research focuses on the FC2 dataset, as it exhibits greater voltage fluctuations compared to FC1. This characteristic not only reflects the performance of the fuel cell under more complex and variable operating conditions but also increases the difficulty and complexity of prediction, thereby providing a more challenging test environment for the model.

To ensure the accuracy of the model in predicting various scenarios that may be encountered in practical applications, the dataset is divided into training and testing sets. At the critical juncture of 500 hours, a decision was made to purposefully divide the training set due to significant data fluctuations observed at this time point. Historically, such volatile nodes are typically avoided by researchers to circumvent potential difficulties in model predictions dealing with unstable data. The model is aimed to identify and adapt to these complex data patterns during training. Conversely, the testing set comprises the subsequent 500 hours of data, which are used to validate the accuracy of the model in handling previously unseen data.

\subsection{Prediction of Temporal Scale Transformer Model}
Prediction experiments were conducted on the ageing dataset. The training and testing protocols remained consistent across the model, yielding the best results. Evaluation metrics, including $Score_{RUL}$, $RMSE$, $A_{FT}$, and $\%Er_{FT}$,  were employed to assess performance. 
The TSTransformer model has demonstrated its robust ability to accurately capture and predict the anomalies at crucial moments, which is vital for enhancing the generalisability and adaptive capacity of predictive systems.

\begin{figure}[ht]
  \centering
  \includegraphics[width=0.5\textwidth]{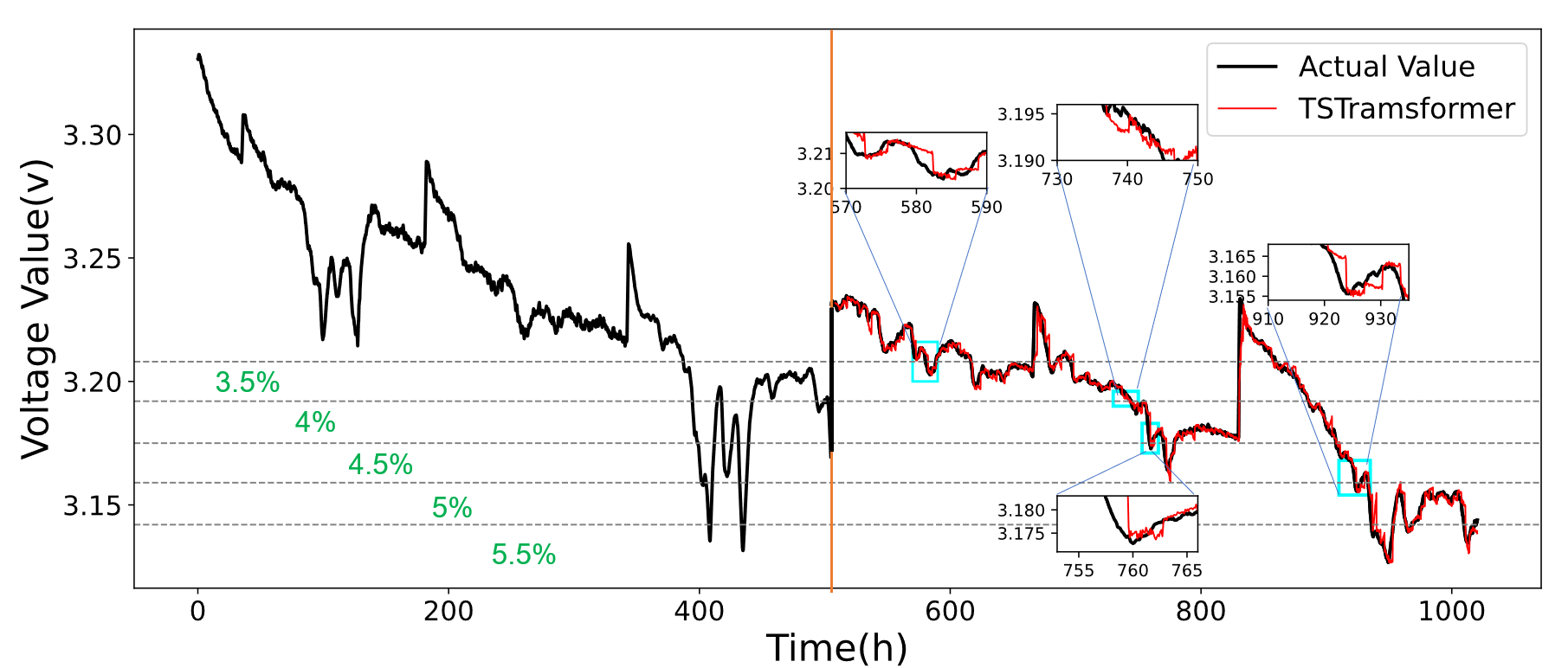}
  \caption{Prediction results of the TSTransformer model.}
  \label{fig:select feature1}
\end{figure}

Fig \ref{fig:select feature1} displays the precise predictive capabilities of the TSTransformer model across different fault thresholds, with the highest $\%Er_{FT}$ not exceeding -2.706\%. Notably, at the 4\% fault threshold, the difference between the predicted and actual RUL is only 0.466\%, showcasing the high accuracy of the model. Table \ref{tab:iTransformer_prediction_results table}. benchmarks the performance of our model against other models, where the TSTransformer excels in the ScoreRUL metric with an RMSE of just 0.0033, significantly outperforming traditional \gls{LSTM} and Transformer models.

\begin{table}[ht]
\centering
\caption{Prediction results of the TSTransformer model on the test set.}
\label{tab:iTransformer_prediction_results table}
\resizebox{0.5\textwidth}{!}{%
\begin{tabular}{cccccc}
\hline
\textbf{Fault threshold}  & \textbf{Actual RUL (h)} & \textbf{Predicted RUL (h)}  & \textbf{\%Er$_{FT}$ } \\
\hline
3.500\% & 80.191  & 82.366 & -2.706\%  \\
4.000\%   & 243.930   & 242.746 & 0.466\% \\
4.500\% & 259.133 & 259.665 & -0.206\%  \\
5.000\%   & 422.751  & 423.893 & -0.269\%  \\
5.500\% & 435.703 & 436.791 & -0.251\% \\
\hline
\end{tabular}%
}
\end{table}

\subsection{The impact of window size on model prediction ability}
By comparing the accuracy of RUL prediction under different window sizes, the overall lag level of the model can be reflected. The degree of lag refers to the time delay between the predicted values of the model and the true values. Specifically, if there is a time difference of \( \delta t \) between the predicted result and the actual value at time \( t \), and the predicted value lags behind, then the model is said to have lag. Lag error can indicate positive prediction advance and negative prediction lag. The specific calculation method can be found in Equation 15. The experiment uses five fault points as the evaluation criteria for testing, and the difference between the actual value and the predicted value at these five fault points can reflect the degree of lag in the model prediction.
     \begin{equation}
     \text{Lag Error} = \hat{y}(t) - y(t + \delta t)
     \end{equation}
Defined as the difference between the predicted value \( \hat{y}(t) \) and the actual value \( y(t + \delta t) \), where \( \delta t \) is the lag time interval. When Lag Error \( < 0 \), it indicates prediction lag; when Lag Error \( > 0 \), it indicates prediction advance.
   
\begin{figure}[ht]
  \centering
  \includegraphics[width=0.5\textwidth]{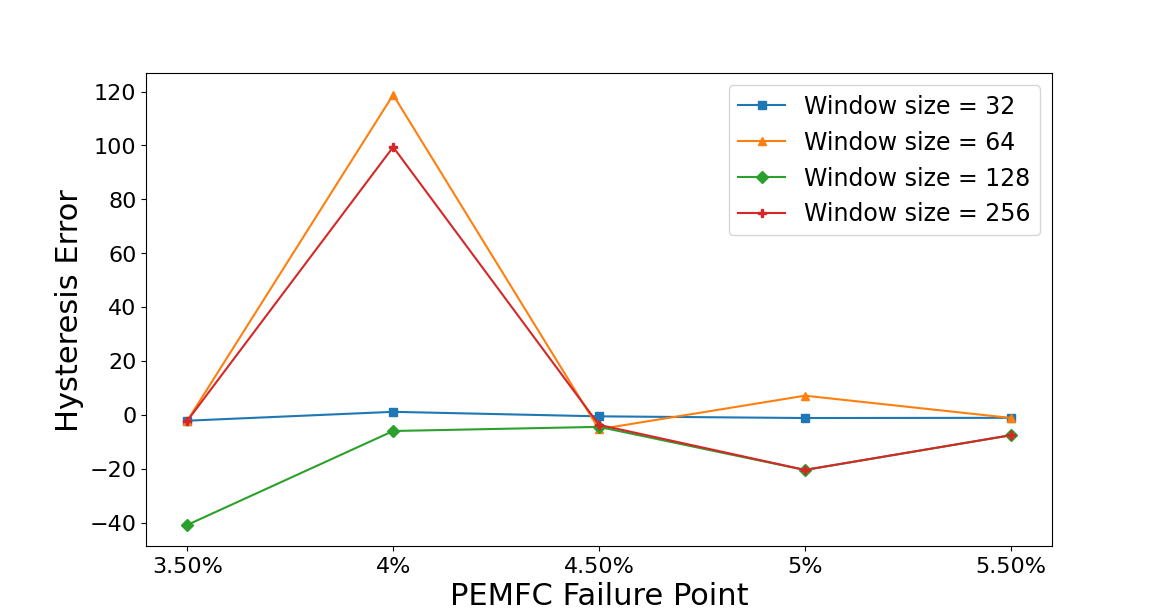}
  \caption{The comparison results of the voltage prediction hysteresis phenomenon under different window sizes.}
  \label{fig:hysteresis}
\end{figure}

Fig \ref{fig:hysteresis} shows that the lag error is minimised at a window size of 32 and fluctuates around zero at all fault points, indicating high prediction accuracy. The window size balances capturing relevant historical information and avoiding external noise that may hurt prediction quality. As the window size increases to 64, a significant peak in lag error at 4\% of the failure points is observed. This may mean that a model with this window size may contain too much historical data, causing it to miss the most recent key changes and resulting in significant prediction delays. It is interesting that the lag error at window size 64 is similar to the lag error at window size 32 except for a 4\% failure point. This indicates that the performance of the model is sensitive to specific failure points at this window size. There is a consistent trend of negative lag error for window sizes of 128 and 256, especially at 5\% and 5.5\% of fault points. This shows that there is a system lag in the prediction of the model relative to the actual occurrence of voltage drop. The negative lag error here means that the model's predictions are delayed, failing to provide early warning, which may be crucial for preemptive maintenance and operational decisions. This trend may result from the model's excessive smoothing of time series data, leading to hysteresis in detecting voltage drop trends.

The trend of increasing lag error with increasing window size indicates a trade-off between capturing long-term dependencies and maintaining responsiveness to recent changes in the TSTransformer model. Although a larger window size allows the model to understand long-term patterns, it may also weaken the influence of recent observations, which are often crucial for timely prediction.

\subsection{Model comparison experiment}

\begin{figure*}[!t]
  \centering
  \includegraphics[width=0.8\textwidth]{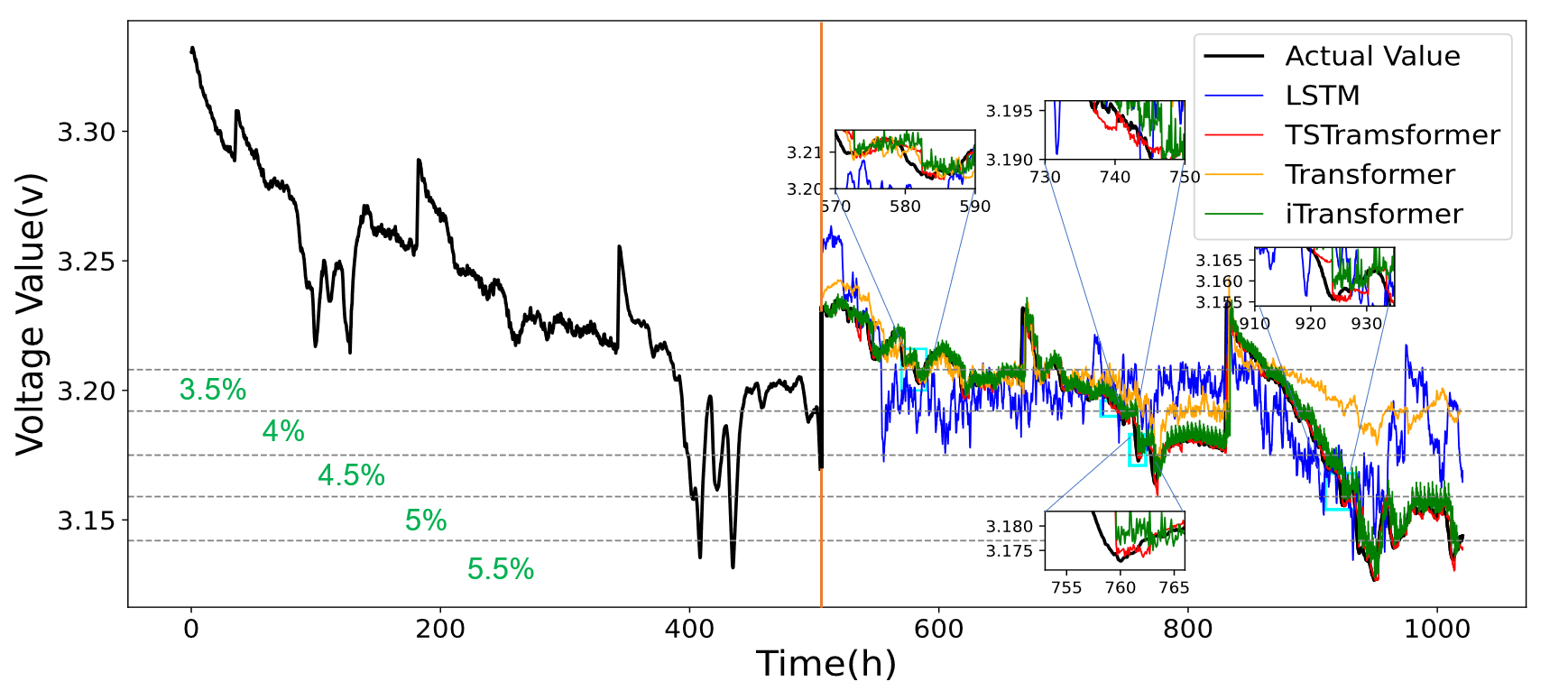}
  \caption{Prediction results of multiple models on the same test set.}
  \label{fig:select feature2}
\end{figure*}

In the case study, a consistent time series dataset was employed to assess the predictive capabilities of the TSTransformer, iTransformer, Transformer, and \gls{LSTM} models in predicting the \gls{PEMFC} degradation. The results revealed that the TSTransformer outperformed the other models in capturing the non-linear features and degradation trends of the \gls{PEMFC}. Particularly noteworthy was the ability of iTransformer to predict future changes when faced with limited training data. This is due to its improved core functionalities, including the reconfiguration and optimisation of model components, which allow better adaptation to the characteristics of time series data. This leads to enhanced predictive performance, particularly under data scarcity.

However, iTransformer encountered challenges capturing long-term trends as the training data size increased. This could impact its prediction accuracy. In contrast, the TSTransformer effectively overcame this challenge by dynamically adjusting the scale of K and V matrices. Especially demonstrating the performance of the TSTransformer in handling high-variability data. This model has proved its effectiveness in medium to long-term and even ultra-long-term prediction of the degradation of the \gls{PEMFC}.

By comparing the performance of different models under identical experimental conditions, the data in Table \ref{tab:Comparsion result with different models} illustrates the superiority of the TSTransformer. Especially in terms of the \gls{RUL} score ($Score_{RUL}$), it achieves a high score of 0.914. This score not only far surpasses the traditional \gls{LSTM} model (0.463) but also significantly outperforms the Transformer (0.013) and iTransformer (0.888). This highlighted score indicates a substantial improvement in predictive accuracy for the TSTransformer. The \gls{RUL} score is a comprehensive metric that reflects the performance of the model across the entire test set, assessing the accuracy and reliability of the model predictions. In this regard, the TSTransformer demonstrates an outstanding performance, affirming its leadership position in similar tasks.

Similarly, the \gls{RMSE} assesses the deviation between model predictions and actual values, serving as another crucial metric. The TSTransformer scores 0.0033 on this metric, which is significantly lower than the other models. This indicates minimal errors in its predictive results and further affirms its performance in predictive accuracy.

\begin{table}[ht]
\centering
\caption{Benchmark of Models}
\label{tab:Comparsion result with different models}
\begin{tabular}{lS[table-format=1.2]S[table-format=1.2]S[table-format=1.2]S[table-format=1.2]}
\toprule
\textbf{Metric} & \textbf{LSTM} & \textbf{Transformer} & \textbf{iTransformer} & \textbf{Proposed} \\
\midrule
$ Score_{R U L}$ & 0.463 & 0.013 & 0.888 & \textbf{0.914} \\
$ RMSE $ & 0.022 & 0.025 & 0.008 & \textbf{0.003} \\
\bottomrule
\end{tabular}
\end{table}

Fig \ref{fig:select feature2} illustrates the predictive accuracy of the TSTransformer at different fault thresholds in detail. Particularly in the enlarged portion of Fig \ref{fig:select feature2}, it can be observed that the predictive curve of the TSTransformer model demonstrates a more pronounced consistency with the actual data at critical moments closely aligned with fault thresholds. For instance, at 4\% voltage loss points, the TSTransformer predicts the imminent voltage drop. At the 4\% fault threshold, the model prediction error with respect to the actual \gls{RUL} is only 0.466\%, a result far superior to other models and highlighting its highly accurate predictive capabilities at critical thresholds.



\section{Conclusion and Future Work}

This study introduces an innovative approach to predict the \gls{RUL} of the \gls{PEMFC} by employing the TSTransformer model, which is a data-driven predictive model architecture. Validated using an ageing dataset, the model adhered to consistent training and testing protocols across evaluations, yielding compelling results. The salient conclusions drawn from this study are:

\begin{enumerate}
    \item The TSTransformer model exhibited robust predictive accuracy, especially at pivotal data fluctuation points, such as the 500-hour threshold. The capability of this model to accurately predict complex nonlinear degradation characteristics of the \gls{PEMFC} stack demonstrates its applicability in dynamic and unpredictable operational conditions.
    
    \item The performance of the model was superior across various fault thresholds, with a maximum error percentage (\%Er\textsubscript{FT}) of -2.706\%. At a critical fault threshold of 4\%, the model achieved an impressive prediction accuracy with a mere 0.466\% deviation from the actual RUL. Furthermore, it surpassed traditional \gls{LSTM} and Transformer models with a Score\textsubscript{RUL} metric and an \gls{RMSE} of only 0.0033, indicating high fidelity in degradation voltage predictions.
    
    \item Comparative analysis revealed that the TSTransformer outperformed iTransformer, Transformer, and LSTM models in terms of capturing intricate details and degradation patterns in the \gls{PEMFC}. Its dynamic scaling ability, adjusting the Key and Value matrices scales, was essential in handling extensive training data and highly variable datasets, rendering it ideal for mid to ultra-long-term predictive tasks.
\end{enumerate}

In essence, the TSTransformer model is distinguished for its capability to circumvent the limitations of traditional machine learning models that necessitate identical distribution in training and test data. Its algorithmic sophistication, tailored to address data fluctuations and degradation trends, positions it as an advanced predictive tool for the RUL prognostication of the PEMFC. The robustness, precision, and adaptability of the model under varied operational conditions mark a significant stride in predictive maintenance for the PEMFC. Future work may include applying this model in real vehicle scenarios to affirm its efficacy and integrate online transfer learning methodologies for real-time application contexts.



\bibliographystyle{ieeetr}
\bibliography{ref}

\vspace{12pt}

\end{document}